\documentclass[11pt,a4paper]{article}
\usepackage[T1]{fontenc}
\usepackage[utf8]{inputenc}
\usepackage{lmodern}
\usepackage[a4paper,margin=1in,headheight=14pt]{geometry}
\usepackage{microtype}
\microtypesetup{expansion=false}
\usepackage{titlesec}
\usepackage{fancyhdr}
\usepackage{graphicx}
\usepackage{booktabs,tabularx,array}
\usepackage{amsmath,amssymb}
\usepackage[font=small,labelfont=bf,labelsep=period]{caption}
\usepackage[numbers,sort&compress]{natbib}
\usepackage{xurl}
\usepackage[unicode,hidelinks]{hyperref}
\usepackage{xcolor}
\graphicspath{{figures/}}
\newcommand{\kcat}{\ensuremath{k_{\mathrm{cat}}}}
\newcommand{\Km}{\ensuremath{K_{\mathrm{m}}}}

\hypersetup{pdftitle={KineCouple: Substrate-conditioned residual readout for enzyme kinetic prediction},pdfauthor={KineCouple Research Team},pdfsubject={Enzyme kinetic prediction}}
\renewcommand{\arraystretch}{1.15}
\titleformat{\section}{\large\bfseries}{\thesection}{0.6em}{}
\titleformat{\subsection}{\normalsize\bfseries}{\thesubsection}{0.6em}{}
\titlespacing*{\section}{0pt}{1.0em}{0.35em}
\titlespacing*{\subsection}{0pt}{0.8em}{0.2em}
\fancypagestyle{plain}{\fancyhf{}\fancyfoot[C]{\small\thepage}}
\title{\vspace{-1.3em}\LARGE\bfseries KineCouple: Substrate-conditioned residual\\readout for enzyme kinetic prediction}
\author{Chun Kang$^{1,2}$, Yunbo Zhang$^{1}$, Fanhu Zeng$^{3}$, Jian Liu$^{1}$, Siwei Deng$^{2}$, Lei Sha$^{1,2}$\thanks{~Corresponding author.}\\
 $^{1}$ School of Artificial Intelligence, Beihang University \\
$^{2}$ OxTium AI Research\\
$^{3}$ Imperial College London}
\date{Preprint\quad October 18, 2026}

\begin{document}
\maketitle

\begin{abstract}
Accurate prediction of enzyme kinetic parameters can support enzyme engineering and metabolic modeling by complementing experimental characterization. Existing prediction methods often pool residue-level protein features before combining them with substrate features. Such substrate-independent pooling cannot adapt residue weighting to the substrate during aggregation, potentially diluting information relevant to a specific enzyme–substrate pair. Since the contributions of individual residues to catalysis can depend on the substrate, substrate-conditioned aggregation offers a way to emphasize residue features relevant to the prediction. We introduce KineCouple, which uses substrate queries and residue-level keys and values in cross-attention, allowing the substrate to guide residue selection before aggregation. The model preserves the global protein representation and adds a gated correction defined as the difference between the conditioned readout and a uniform readout. ProtT5 and MolT5 remain frozen, and separate models predict \kcat{} and \Km{} on a logarithmic scale. In pooled out-of-fold predictions on the CataPro benchmarks, KineCouple reduces \kcat{} RMSE from 1.333 to 1.330 and increases PCC from 0.492 to 0.496 relative to reproduced CataPro. For \Km{}, RMSE decreases from 0.998 to 0.997 and PCC increases from 0.633 to 0.634. Spearman correlation also improves for both tasks. 
\end{abstract}

\section{Introduction}
\label{sec:introduction}
Enzyme kinetic parameters provide a quantitative basis for understanding catalytic behavior. The catalytic turnover number \kcat{} describes the turnover rate per catalytic site at saturating substrate concentrations, whereas the Michaelis constant \Km{} denotes the substrate concentration at which the reaction rate reaches half its maximum under Michaelis--Menten kinetics. These parameters are important for comparing enzyme variants, guiding enzyme engineering, and constructing kinetic models of metabolic pathways. However, their experimental determination is labor-intensive, and the measured values depend on the enzyme, substrate, and assay conditions, including temperature and pH. Computational prediction can complement experimental characterization by estimating kinetic parameters for enzyme--substrate pairs lacking measurements. Recent methods, including UniKP and CataPro, use pretrained representations of protein sequences and substrate structures for this task~\citep{yu2023unikp,wang2025robust}.

Existing methods such as UniKP and CataPro encode enzyme sequences and substrate structures separately, then combine their representations to predict kinetic parameters. In these methods, residue-level protein features are pooled into a global representation before fusion with substrate features. Although a downstream predictor can learn interactions between the resulting representations, the protein aggregation itself remains independent of the substrate. This raises the question of whether incorporating substrate context into enzyme representations can improve kinetic prediction.

We propose KineCouple to learn substrate-aware enzyme representations while keeping the pretrained encoders frozen. Cross-attention allows substrate features to guide residue weighting before aggregation, providing a mechanism for integrating complementary information from enzyme sequences and substrate structures. This design aims to capture features relevant to each enzyme--substrate pair while retaining global protein context. A gated correction derived from the difference between the substrate-conditioned and uniform readouts is added to the global protein vector. The updated protein representation is then combined with the original substrate features for kinetic prediction.

We train separate models for \kcat{} and \Km{} and compare their pooled out-of-fold (OOF) predictions with reproduced CataPro. RMSE decreases and both Pearson and Spearman correlations increase for both tasks, with absolute differences of 0.001--0.004 at the reported precision. The contributions are twofold:
\begin{enumerate}
 \item \textbf{Substrate-guided selection of residue information.} Conditioning the protein readout on the substrate allows the model to assign residue weights for each enzyme--substrate pair while retaining global protein information. The aim is to emphasize features relevant to the current prediction, without requiring predefined active-site annotations.
 \item \textbf{Finer-grained interaction through cross-attention.} Substrate-derived queries interact with residue-derived keys and values before protein aggregation, extending interaction beyond pooled global vectors. A centered residual and a gate connect the resulting correction to the original prediction pathway.
\end{enumerate}

\section{Related work}
\label{sec:related}
\subsection{Predicting enzyme kinetic parameters}
Predicting kinetic parameters from enzyme sequences and substrate descriptors offers a way to estimate parameters where kinetic measurements are scarce. DLKcat combines convolutional sequence processing with a molecular graph encoder for turnover prediction and enzyme-constrained metabolic modeling~\citep{li2022deep}. UniKP uses pretrained protein and substrate embeddings to predict several kinetic parameters~\citep{yu2023unikp}. CataPro combines ProtT5, MolT5, and molecular fingerprints and evaluates predictions using sequence-similarity-aware data splits~\citep{wang2025robust}. CatPred addresses kinetic dataset construction, representation combinations, and predictive uncertainty~\citep{boorla2025catpred}. These approaches provide estimates for enzyme--substrate pairs, with performance depending on the available inputs and evaluation setting.

Other studies extend the inputs or the prediction task. TurNuP incorporates substrates and products into reaction fingerprints and examines generalization to enzymes and reactions distinct from the training data~\citep{kroll2023turnover}. EITLEM-Kinetics uses ensemble iterative transfer learning to predict kinetic parameters of enzyme mutants~\citep{shen2024eitlem}, whereas DLTKcat explicitly models temperature-dependent turnover rates~\citep{qiu2024dltkcat}. Full reactions, mutation effects, and assay conditions impose different information requirements; a single aggregate metric cannot substitute for evaluation of each setting. Here, we retain sequence--substrate inputs and absolute-parameter supervision, focusing on the readout mechanism. Reproduced CataPro, with the same global features and predictor architecture, serves as the direct baseline.

\subsection{Protein--substrate representations and residue-level interaction}
Pretrained models provide reusable protein and molecular features for kinetic prediction. ProtT5, introduced in ProtTrans, produces transferable contextual residue representations. ESM-2 demonstrates that large sequence language models learn representations containing protein structural information~\citep{lin2023evolutionary}. MolT5 learns representations for molecule--text translation through pretraining on molecular strings and natural language~\citep{edwards2022translation}. Pooling these features before fusion supports learning relationships between global vectors, but prevents the substrate from guiding residue selection. Introducing conditional interaction while retaining pretrained features is therefore a modeling choice distinct from encoder selection.

Previous methods address residue interaction and structural information in several ways. PMAK introduces residue-aware attention between enzyme residues and reaction representations~\citep{cai2026enhancing}. DeepEnzyme combines sequence representations with protein three-dimensional structural graph features to predict turnover~\citep{wang2024deepenzyme}. SaProt integrates amino acid identities and discrete local structural tokens into a joint pretraining vocabulary~\citep{su2024saprot}. ERBA uses substrate-conditioned cross-attention during protein language model adaptation and incorporates active-site geometry through a mixture-of-experts module~\citep{wang2026multimodal}. Thus, neither attention nor the use of residue features alone is a new contribution of this work.

KineCouple places a readout outside the frozen encoders: a whole-substrate vector queries residue features, and the change relative to a uniform value readout becomes a correction to the global protein representation. Here, finer-grained interaction refers to the residue level rather than interaction solely between pooled vectors; the model does not construct a substrate-atom--residue contact matrix. We evaluate this combined readout and fusion design. Published results from the methods above are not treated as reproductions under a common OOF protocol.

\section{Method}
\label{sec:method}
\subsection{Overview}
KineCouple takes an enzyme sequence and substrate SMILES as input and predicts either \kcat{} or \Km{} on a logarithmic scale. Frozen ProtT5 produces residue features $H$, whose mean over valid positions defines the global protein vector $p_0$. A frozen MolT5 representation $t$ is concatenated with a MACCS fingerprint $f$ to form substrate features $s=[t;f]$. Neither encoder is updated during task training.

The model retains $p_0$ as a global pathway and uses $s$ to guide residue readout. The resulting correction is fused with $p_0$, and the updated representation enters the predictor alongside the original substrate features. Figure~\ref{fig:architecture} shows the information flow; dimensions and predictor details are given in Appendix~\ref{app:implementation}.

\begin{figure}[t]
  \centering
  \IfFileExists{figures/overall_3d.png}{%
    \includegraphics[width=\linewidth]{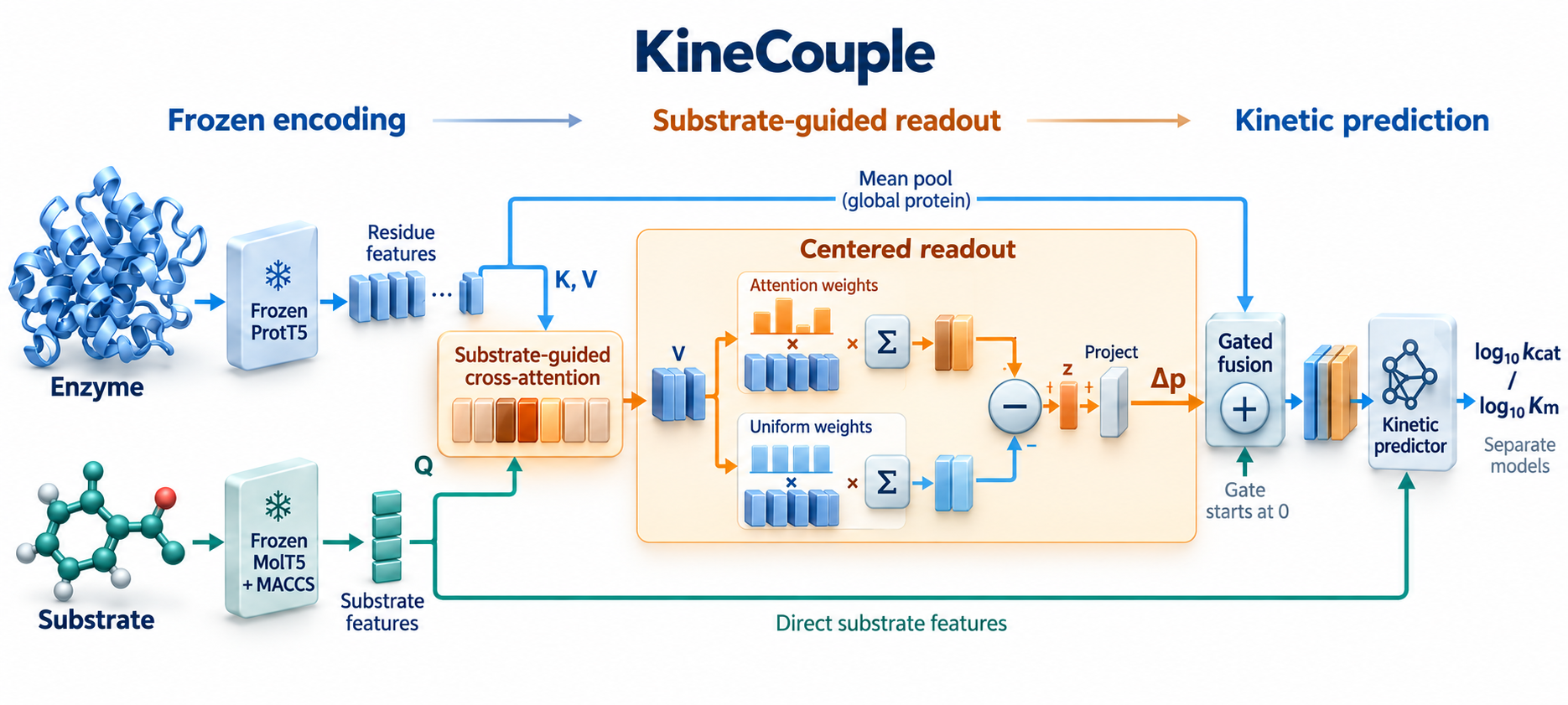}%
  }{%
    \fbox{\parbox[c][45mm][c]{0.94\linewidth}{\centering\small
    Architecture figure is provided as `figures/overall_3d.png` in the full project.\\
    The single-file editor preview is compiled without auxiliary figure files.}}
  }
  \caption{\textbf{Substrate-conditioned centered residual readout.} Frozen ProtT5 produces a residue matrix $H$ and a global mean $p_0$; frozen MolT5 and MACCS produce substrate features $t$, $f$, and $s=[t;f]$. The substrate conditioner generates queries for masked multi-head cross-attention over residue keys and values. The centered readout subtracts uniform value pooling, maps the result to $\Delta p$, and injects it into $p_0$ through $\tanh(g)$. The enhanced protein vector and direct substrate features enter the enzyme--substrate fusion predictor.}
  \label{fig:architecture}
\end{figure}

\subsection{Residue selection through cross-attention}
A substrate conditioning network converts $s$ into a query $q$. Inspired by GeneLLM~\cite{deng2026decoding}, residue features are normalized and independently projected to keys $k_i$ and values $v_i$. The query determines how residues are read: positions whose keys better match the current substrate query receive larger weights, and the value vectors supply the information to aggregate. For head $h$, the residue weights are
\begin{equation}
 \alpha_{hi}=\operatorname{softmax}_{i\in\mathcal I}
 \left(\frac{q_h^{\mathsf T}k_{hi}}{\sqrt{d_h}}\right),
 \label{eq:attention}
\end{equation}
where $\mathcal I$ is the set of valid residues. We use four attention heads with $d_h=32$; padding positions are excluded from normalization.

For a fixed model and enzyme, residue keys and values remain unchanged across substrates, while queries and attention weights can vary. The substrate can therefore influence residue selection before protein aggregation, in addition to serving as an input to the final regression. These weights describe the model's use of features and do not directly identify physical binding sites.

\subsection{Centered readout and gated fusion}
To retain the original global pathway, the conditional branch represents the change introduced by reweighting. For each head, we subtract the uniform mean in the same value space from the attention-weighted readout:
\begin{equation}
 z_h=\sum_{i\in\mathcal I}\left(\alpha_{hi}-\frac{1}{n}\right)v_{hi},
 \qquad n=|\mathcal I|.
 \label{eq:centered}
\end{equation}
Uniform attention gives a zero difference. The four head outputs are concatenated and projected without bias to obtain $\Delta p$. Gated fusion and prediction then take the form
\begin{equation}
 p^*=p_0+\tanh(g)\Delta p,\qquad \widehat y=F(p^*,s).
 \label{eq:gate}
\end{equation}
The scalar $g$ is shared within each task and initialized at zero. With matching predictor weights, inputs, and evaluation state, the initial model recovers the base prediction function. Training then learns both the correction and its injection strength. Centering occurs in the projected value space; it does not subtract the original protein vector $p_0$.

\paragraph{Training.}
We first train the global base model, then load its weights and train the added branch together with the original prediction head. ProtT5 and MolT5 remain frozen. The \kcat{} and \Km{} models have independent trainable parameters and share only frozen feature caches. Training uses mean squared error scaled by the training-fold label standard deviation; network outputs remain on the original logarithmic scale. Appendix~\ref{app:implementation} specifies the conditioning network, initialization gradients, and loss.

\section{Experiments}
\label{sec:experiments}
\subsection{Evaluation setting}
The primary comparison uses the CataPro \kcat{} and \Km{} benchmarks. The reproduced baseline and KineCouple correspond to the pooled OOF results labeled \texttt{base} and \texttt{attention} in the experiment record. Metrics are computed after merging held-out predictions across folds, rather than averaged across random seeds. The available record does not include fold assignments, sample counts, checkpoint selection rules, or run configurations, so complete agreement between the two evaluation protocols has not been verified.

We report root mean squared error (RMSE), Pearson correlation (PCC), and Spearman correlation (SCC). The logarithmic targets are described in Section~\ref{sec:method}. The comparison record reports three decimal places but does not provide raw predictions, unit-conversion metadata, confidence intervals, or repeated-run statistics. We therefore report metric differences at the available precision without significance testing.

\subsection{Main method comparison}
Table~\ref{tab:main-results} includes five methods: DLKcat, UniKP, reproduced CataPro, CataPro+SaProt, and KineCouple. DLKcat and UniKP represent sequence--substrate prediction approaches, reproduced CataPro is the direct baseline, and CataPro+SaProt provides a structure-enhanced reference. All values come from the existing comparison record. Reproduced CataPro and KineCouple use pooled OOF results; the other three rows contain reference-table values. The five methods have not been rerun under a common protocol.

Among these reported values, KineCouple has the highest PCC and SCC for both tasks and the lowest \Km{} RMSE; CataPro+SaProt has the lowest \kcat{} RMSE. Relative to reproduced CataPro, KineCouple reduces \kcat{} RMSE from 1.327 to 1.319 and increases PCC and SCC from 0.481 to 0.496 and from 0.470 to 0.495, respectively. For \Km{}, RMSE decreases from 1.006 to 0.997, PCC increases from 0.622 to 0.634, and SCC increases from 0.610 to 0.631. All six changes favor KineCouple, but whether they exceed variation between training runs remains unknown.

\begin{table}[t]
\centering
\caption{\textbf{Comparison of five methods for kinetic prediction.} Bold indicates the best reported value in each column. $\dagger$: pooled OOF results; $\ddagger$: reference-table values. Split and aggregation protocols have not been verified across sources, and variability across random seeds is unavailable.}
\label{tab:main-results}
\small
\setlength{\tabcolsep}{4pt}
\begin{tabular*}{\linewidth}{@{\extracolsep{\fill}}lrrrrrr@{}}
\toprule
 & \multicolumn{3}{c}{\kcat{}} & \multicolumn{3}{c}{\Km{}}\\
\cmidrule(lr){2-4}\cmidrule(l){5-7}
Method & RMSE $\downarrow$ & PCC $\uparrow$ & SCC $\uparrow$ & RMSE $\downarrow$ & PCC $\uparrow$ & SCC $\uparrow$\\
\midrule
DLKcat$^{\ddagger}$ & 1.426 & 0.368 & 0.349 & 1.052 & 0.578 & 0.578\\
UniKP$^{\ddagger}$ & 1.382 & 0.448 & 0.449 & 1.039 & 0.599 & 0.594\\
CataPro (reproduced)$^{\dagger}$ & 1.327 & 0.481 & 0.470 & 1.006 & 0.622 & 0.610\\
CataPro+SaProt$^{\ddagger}$ & \textbf{1.317} & 0.493 & 0.489 & 0.999 & 0.631 & 0.625\\
KineCouple$^{\dagger}$ & 1.319 & \textbf{0.496} & \textbf{0.495} & \textbf{0.997} & \textbf{0.634} & \textbf{0.631}\\
\bottomrule
\end{tabular*}
\end{table}

\subsection{Extended reference comparison}
Appendix~\ref{app:reference} lists all 18 models in the comparison record. Sixteen rows come from a reference table, while reproduced CataPro and KineCouple use the pooled OOF results described above. Among the listed values, KineCouple has the highest PCC and SCC for both tasks and the lowest \Km{} RMSE. CataPro+SaProt has a lower \kcat{} RMSE of 1.317, compared with 1.319 for KineCouple. These rankings are descriptive because splits, preprocessing, and aggregation procedures have not been reconciled across sources.

\subsection{Ablation Study}
\label{sec:ablation}

To investigate the contributions of different components in
KineCouple, we conduct ablation studies on the $k_{\mathrm{cat}}$
and $K_m$ prediction benchmarks. We report the average results
over ten Folds, with three random seeds per fold.
Five variants are considered:
(1) \textbf{Baseline (A)}, the original CataPro model;
(2) \textbf{Continued Training (B)}, which further optimizes
the original prediction head to control for additional training;
(3) \textbf{Mean Readout (P)}, which aggregates residue features
through mean pooling;
(4) \textbf{Fixed-Query Attention (Q)}, which employs a learned,
substrate-independent query for residue-level attention;
and (5) \textbf{Full Model (F)}, which incorporates
substrate-conditioned attention and centered gated residual
readout. All variants share the same data splits and frozen
encoder features. Additional implementation details are
provided in the Appendix.

\begin{table*}[t]
\centering
\caption{
Ablation study of KineCouple on the $k_{\mathrm{cat}}$
and $K_m$ prediction benchmarks.
Results are averaged over Folds 1, 5, 6, and 9,
with three random seeds per fold.
The best results are highlighted in bold.
Lower RMSE and MAE indicate better performance,
whereas higher PCC, SCC, and $R^2$ are preferred.
}
\label{tab:ablation}
\setlength{\tabcolsep}{5pt}
\renewcommand{\arraystretch}{1.15}
\resizebox{\textwidth}{!}{
\begin{tabular}{lcccccc}
\toprule
& \multicolumn{3}{c}{$k_{\mathrm{cat}}$}
& \multicolumn{3}{c}{$K_m$} \\
\cmidrule(lr){2-4}
\cmidrule(lr){5-7}
\textbf{Variant}
& RMSE $\downarrow$
& PCC $\uparrow$
& SCC $\uparrow$
& RMSE $\downarrow$
& PCC $\uparrow$
& SCC $\uparrow$\\
\midrule
Baseline (A)
& 1.3274 & 0.4808 & 0.4699
& 1.0056 & 0.6217 & 0.6103\\

Continued Training (B)
& 1.3238 & 0.4855 & 0.4725
& 1.0039 & 0.6239 & 0.6148 \\

Mean Readout (P)
& 1.3204 & 0.4882 & 0.4748
& 1.0032 & 0.6245
& 0.6150\\

Fixed-Query Attention (Q)
& 1.3188 & 0.4899
& 0.4751
& 1.0039 & 0.6238
& 0.6152\\

\midrule
\textbf{KineCouple (F)}
& \textbf{1.3185}
& \textbf{0.4962} & \textbf{0.4948}
& \textbf{0.9974} & \textbf{0.6344} & \textbf{0.6309}\\
\bottomrule
\end{tabular}
}
\end{table*}

As shown in Table~\ref{tab:ablation}, KineCouple achieves
the best performance across all six evaluation metrics
on both $k_{\mathrm{cat}}$ and $K_m$ prediction tasks.
For $k_{\mathrm{cat}}$ prediction, the full model achieves
an RMSE of 1.3185, a PCC of 0.4962, and an SCC of 0.4948.
Compared with the original CataPro baseline, KineCouple
reduces RMSE by 0.67\%, while improving PCC and SCC
by 0.0154 and 0.0249, respectively.
For $K_m$ prediction, KineCouple reduces RMSE from
1.0056 to 0.9974, corresponding to a relative reduction
of 0.82\%, while increasing PCC from 0.6217 to 0.6344
and SCC from 0.6103 to 0.6309.
Furthermore, the full model consistently outperforms
continued training (B), mean readout (P), and fixed-query
attention (Q), demonstrating improvements in both
prediction accuracy and correlation with measured
kinetic parameters.

The ablation results provide insights into the contributions
of the proposed components.
First, continued training improves performance over the
original baseline, indicating that additional optimization
accounts for part of the observed gains.
Introducing residue-level readout further improves
prediction accuracy. In particular, for $k_{\mathrm{cat}}$,
RMSE decreases from 1.3238 with continued training
to 1.3204 with mean readout, and further to 1.3188
with fixed-query attention, suggesting the benefit of
residue-level feature aggregation and attention-based
selection.
More importantly, replacing the fixed query with
a substrate-conditioned query leads to additional
improvements across both tasks.
Compared with fixed-query attention, KineCouple improves
PCC from 0.4899 to 0.4962 and SCC from 0.4751 to 0.4948
for $k_{\mathrm{cat}}$ prediction.
For $K_m$, RMSE decreases from 1.0039 to 0.9974,
while PCC and SCC increase by 0.0106 and 0.0157,
respectively.
These findings suggest that substrate-dependent residue
selection provides additional predictive information
beyond unconditional attention-based aggregation.
Overall, the results support the effectiveness of
substrate-conditioned residue readout for modeling
fine-grained enzyme-substrate interactions, with
consistent improvements across both kinetic
prediction tasks.

\section{Discussion and limitations}
\label{sec:discussion}
The reported results favor KineCouple, but the comparison changes several factors simultaneously. In addition to conditional attention, KineCouple adds 533,249 trainable parameters and continues training the original prediction head. A base-model continuation control with the same training budget is needed to distinguish further optimization from the effect of the new branch. Uncentered and unconditional readout controls are also needed to test centering and substrate dependence. The current results do not rule out these alternative explanations.

The small aggregate differences require more complete statistical analysis. Raw OOF predictions would support paired error comparisons, while repeated training would quantify sensitivity to initialization and optimization; neither is available in the current record. Higher aggregate correlations also do not establish better prediction of differences between substrates for the same enzyme or of mutation responses. These questions require the matched, paired evaluations described in Appendix~\ref{app:validation}.

The model is also limited by its inputs. It does not explicitly receive pH or temperature, and the sequence-length limit removes middle positions from sequences longer than 1000 residues. Attention weights describe how the predictor combines retained features; they do not establish residue contacts or catalytic roles. Frozen encoders enable feature reuse, but residue caching and projection introduce additional costs that have not yet been measured against the base model.

\section{Conclusion}
\label{sec:conclusion}
KineCouple defines a substrate-conditioned residual relative to uniform value pooling and adds it to a global protein representation. A zero-initialized gate preserves the loaded base prediction function, while subsequent training allows the added branch to use residue features. The reported OOF metrics are slightly better than reproduced CataPro on both \kcat{} and \Km{}. Repeated runs and matched ablations are needed to establish the reliability and source of these differences.

\section*{Acknowledgement}

This work was supported by the New Generation Artificial Intelligence-National Science and Technology Major Project (No. 2026ZD0128000, No. 2026ZD0128003).

\section*{Data and code availability}
Dataset version identifiers, fold assignments, raw OOF predictions, training configurations, and a versioned code repository have not yet been provided. Experimental results are currently available only as aggregate metrics rounded to three decimal places.

\section*{Declarations}
Author names, affiliations, correspondence, author contributions, funding, and competing-interest statements remain to be supplied and confirmed by the authors.

\bibliographystyle{unsrtnat}
\bibliography{references}

\newpage
\appendix
\section{Implementation details}
\label{app:implementation}
\paragraph{Feature preprocessing.}
Sequences of at most 1000 residues are retained in full. For longer sequences, only the first 500 and last 500 residues are retained. Pooling and attention exclude padding and removed positions. Encoders operate in evaluation mode during feature extraction, and pooling excludes special tokens and padding. The implementation checks that the mean of the retained residue features agrees with the cached global protein vector.

\paragraph{Conditioning network and projections.}
The valid-residue mean is $p_0=n^{-1}\sum_{i\in\mathcal I}H_i$. The conditioning network normalizes the MolT5 vector, concatenates it with the MACCS fingerprint, and applies a linear layer, SiLU, and LayerNorm:
\begin{equation}
 c=\operatorname{LN}_c\!\left(\operatorname{SiLU}\!\left(W_c[\operatorname{LN}_t(t);f]+b_c\right)\right)\in\mathbb R^{128}.
 \label{eq:condition}
\end{equation}
Queries, keys, and values are given by
\begin{equation}
 q=W_qc,\qquad k_i=W_k\operatorname{LN}_H(H_i),\qquad v_i=W_v\operatorname{LN}_H(H_i).
 \label{eq:qkv}
\end{equation}
Let $\bar v_h=n^{-1}\sum_{i\in\mathcal I}v_{hi}$. Since attention weights sum to one,
\begin{equation}
 z_h=\sum_{i\in\mathcal I}\alpha_{hi}(v_{hi}-\bar v_h),\qquad
 \Delta p=W_o\operatorname{Concat}(z_1,z_2,z_3,z_4).
 \label{eq:centered_equiv}
\end{equation}
The conditional branch thus combines deviations of projected features from their mean. The correction is not constrained to be orthogonal to $p_0$.

\paragraph{Training objective and initialization.}
For task $\tau$, the loss is
\begin{equation}
 \mathcal L_\tau=\frac{1}{B}\sum_{b=1}^{B}
 \left(\frac{\widehat y_{\tau,b}-y_{\tau,b}}{\sigma_{\tau,\mathrm{train}}}\right)^2.
 \label{eq:loss}
\end{equation}
The standard deviation changes only the loss scale. The output projection is randomly initialized. At $g=0$, the internal attention parameters receive zero task gradient, whereas the gate can receive a nonzero gradient through $\Delta p$.

\paragraph{Fusion predictor.}
The protein correction is added before the protein BatchNorm layer. The predictor input is
\begin{equation}
 x=[\operatorname{BN}_{1024}(p^*);\operatorname{BN}_{768}(t);f]\in\mathbb R^{1959},
\end{equation}
followed by a $1959\!\rightarrow\!256$ linear layer, BatchNorm, Dropout, ReLU, and a $256\!\rightarrow\!1$ output layer. Dropout is 0 for \kcat{} and 0.1 for \Km{}. Excluding frozen encoders, the model contains 1,039,362 trainable parameters. Attention probabilities use no dropout, and $W_o$ has no bias, preserving the zero correction under uniform attention.

\paragraph{Output units.}
The scalar predictions correspond to
\begin{equation}
 \widehat y_{\kcat}=\log_{10}\!\left(\widehat{\kcat}/[\mathrm{s}^{-1}]\right),\qquad
 \widehat y_{\Km}=\log_{10}\!\left(\widehat{\Km}/[\mathrm M]\right).
 \label{eq:output}
\end{equation}
The inverse transformation is applied only when reporting predictions in physical units.

\begin{table}[t]
\centering
\caption{Tensor dimensions for one enzyme--substrate pair.}
\small
\begin{tabularx}{\linewidth}{@{}lXl@{}}
\toprule
Variable & Definition & Shape \\
\midrule
$H$ & Frozen ProtT5 residue representations & $L\times1024$ \\
$p_0$ & Mean over valid residues & $1024$ \\
$t$ & Frozen MolT5 substrate vector & $768$ \\
$f$ & MACCS fingerprint & $167$ \\
$s=[t;f]$ & Substrate features supplied to the predictor & $935$ \\
$c$ & Output of the substrate conditioning network & $128$ \\
$z$ & Concatenated centered readouts from four heads & $128$ \\
$\Delta p$ & Correction from the bias-free output projection & $1024$ \\
$x$ & Fusion predictor input & $1959$ \\
$\hat y$ & Log-scale kinetic prediction & $1$ \\
\bottomrule
\end{tabularx}
\end{table}

\section{Extended reference comparison}
\label{app:reference}
Table~\ref{tab:reference-comparison} combines reference-table entries with two sets of pooled OOF results. Model identifiers and numerical precision follow the original comparison record. A shared evaluation protocol across the two sources has not been verified.

\begin{table}[t]
\centering
\caption{\textbf{Extended benchmark comparison.} Lower RMSE and higher PCC and SCC are better. Values are reported to three decimal places; bold indicates the best reported value in each column. $\dagger$: pooled OOF results (\texttt{base} and \texttt{attention}); $\ddagger$: reference-table values. Evaluation protocols may differ across sources.}
\label{tab:reference-comparison}
\small
\setlength{\tabcolsep}{4pt}
\begin{tabular*}{\linewidth}{@{\extracolsep{\fill}}lrrrrrr@{}}
\toprule
 & \multicolumn{3}{c}{\kcat{}} & \multicolumn{3}{c}{\Km{}} \\
\cmidrule(lr){2-4}\cmidrule(l){5-7}
Method & RMSE & PCC & SCC & RMSE & PCC & SCC \\
\midrule
DLKcat$^{\ddagger}$ & 1.426 & 0.368 & 0.349 & 1.052 & 0.578 & 0.578 \\
UniKP$^{\ddagger}$ & 1.382 & 0.448 & 0.449 & 1.039 & 0.599 & 0.594 \\
ESM2\_15B$^{\ddagger}$ & 1.431 & 0.435 & 0.467 & 1.063 & 0.581 & 0.567 \\
ESM2\_RDKitFP$^{\ddagger}$ & 1.380 & 0.435 & 0.424 & 1.025 & 0.609 & 0.607 \\
ESM2\_Morgan$^{\ddagger}$ & 1.370 & 0.446 & 0.436 & 1.017 & 0.617 & 0.616 \\
ESM2\_MACC$^{\ddagger}$ & 1.356 & 0.463 & 0.453 & 1.007 & 0.625 & 0.622 \\
ESM2\_MolT5$^{\ddagger}$ & 1.361 & 0.455 & 0.445 & 1.011 & 0.620 & 0.616 \\
ESM2\_MolT5-MACC$^{\ddagger}$ & 1.357 & 0.464 & 0.459 & 1.004 & 0.627 & 0.623 \\
ProtT5\_3B$^{\ddagger}$ & 1.441 & 0.443 & 0.434 & 1.068 & 0.603 & 0.519 \\
ProtT5\_RDKitFP$^{\ddagger}$ & 1.389 & 0.421 & 0.415 & 1.029 & 0.604 & 0.602 \\
ProtT5\_Morgan$^{\ddagger}$ & 1.364 & 0.465 & 0.462 & 1.015 & 0.621 & 0.620 \\
ProtT5\_Mole-BERT$^{\ddagger}$ & 1.341 & 0.483 & 0.481 & 1.014 & 0.617 & 0.611 \\
ProtT5\_MACC$^{\ddagger}$ & 1.338 & 0.486 & 0.482 & 1.011 & 0.621 & 0.616 \\
ProtT5\_MolT5$^{\ddagger}$ & 1.338 & 0.491 & 0.484 & 1.001 & 0.629 & 0.625 \\
ProtT5-SaProt\_MolT5$^{\ddagger}$ & 1.323 & 0.490 & 0.489 & 1.004 & 0.626 & 0.620 \\
CataPro (reproduced)$^{\dagger}$ & 1.327 & 0.481 & 0.470 &  1.006 & 0.622 & 0.610\\
\midrule
CataPro+SaProt$^{\ddagger}$ & \textbf{1.317} & 0.493 & 0.489 & 0.999 & 0.631 & 0.625 \\
\textbf{KineCouple (ours)}$^{\dagger}$ & 1.330 & \textbf{0.496} & \textbf{0.495} & \textbf{0.997} & \textbf{0.634} & \textbf{0.631} \\
\bottomrule
\end{tabular*}
\end{table}

\clearpage
\section{Planned follow-up analyses}
\label{app:validation}
The current comparison does not report the following controls. Each tests a possible source of the aggregate improvement:
\begin{itemize}
 \item Continue training the base prediction head with the same second-stage training budget to test the effect of additional optimization.
 \item Add a global branch with matched parameter capacity to test the effect of additional trainable parameters.
 \item Remove subtraction of the uniform value readout while retaining the other components to test centering.
 \item Replace the substrate query with a learned unconditional query, or shuffle substrate inputs only within the attention branch, to test substrate conditioning. Shuffling must preserve the predictor's direct substrate input. The unconditional query changes the parameter count; shuffling experiments should be evaluated across multiple fixed random permutations.
\end{itemize}

Evaluating substrate-dependent mutation responses requires matched measurements. For a wild type $W$, mutant $M$, and substrates $A,B$, define
\begin{equation}
 C=[y(M,A)-y(W,A)]-[y(M,B)-y(W,B)].
 \label{eq:contrast}
\end{equation}
Comparing predicted and measured $C$ would test whether the model captures changes in the effect of the same mutation across substrates. The current model is trained on absolute labels, and no evaluation of this contrast is reported here.

\end{document}